\documentclass[5p,times,twocolumn]{elsarticle}

\usepackage[utf8]{inputenc}
\usepackage{amsmath,amssymb,amsfonts,amsthm}
\usepackage[table,xcdraw,dvipsnames,svgnames]{xcolor}
\usepackage{url}
\usepackage{csvsimple}
\usepackage{adjustbox}
\usepackage{hyperref}
\usepackage{pdfpages}
\usepackage{subcaption}
\usepackage{tabularx}
\usepackage{multirow}

\usepackage{algorithmic}
\usepackage{url}
\usepackage{graphicx}
\usepackage{subcaption}
\usepackage{textcomp}
\usepackage[utf8]{inputenc}
\usepackage{dsfont}
\usepackage{hyperref} 
\usepackage{booktabs}

\usepackage[misc,geometry]{ifsym} 
\usepackage[ruled,vlined]{algorithm2e}

\usepackage{multicol}

\usepackage{natbib}
\journal{Information Fusion}

\theoremstyle{definition}

\makeatletter
\def\ps@pprintTitle{%
 \let\@oddhead\@empty
 \let\@evenhead\@empty
 \def\@oddfoot{}%
 \let\@evenfoot\@oddfoot}
\makeatother

\begin{document}

\begin{frontmatter}

\title{Aggregation as Unsupervised Learning and its Evaluation} 

\author[lnu,rise]{Maria Ulan\corref{cor1}}
\ead{maria.ulan@ri.se}

\author[lnu]{Welf L\"owe}
\ead{welf.lowe@lnu.se}

\author[lnu]{Morgan Ericsson}
\ead{morgan.ericsson@lnu.se}

\author[lnu]{Anna Wingkvist}
\ead{anna.wingkvist@lnu.se}

\cortext[cor1]{Corresponding author}
\address[lnu]{Data-driven Software and Information Quality Group,
             Centre for Data Intensive Sciences and Applications,
             Linnaeus University, 351 95 V\"axj\"o, Sweden \\}
\address[rise]{RISE Research Institutes of Sweden, 
               Lindholmspiren 3A, 417 56 G\"othenburg, Sweden\\}

\begin{abstract}
Regression uses supervised machine learning to find a model that combines several independent variables to predict a dependent variable based on ground truth (labeled) data, i.e., tuples of independent and dependent variables (labels). Similarly, aggregation also combines several independent variables to a dependent variable. The dependent variable should preserve properties of the independent variables, e.g., the ranking or relative distance of the independent variable tuples, and/or represent a latent ground truth that is a function of these independent variables. However, ground truth data is not available for finding the aggregation model. Consequently, aggregation models are data agnostic or can only be derived with unsupervised machine learning approaches.

We introduce a novel unsupervised aggregation approach based on intrinsic properties of unlabeled training data, such as the cumulative probability distributions of the single independent variables and their mutual dependencies. 

For assessing this against other aggregation approaches, two perspectives are relevant: (i) how well the aggregation output represents properties of the input tuples, and (ii) how well can aggregated output predict a latent ground truth. We present an empirical evaluation framework that allows to evaluate aggregation approaches from both perspectives. To this end, we use data sets for assessing supervised regression approaches that contain explicit ground truth labels. However, the ground truth is not used for deriving the aggregation models, but it allows for the assessment from a perspective (ii). More specifically, we use regression data sets from the UCI machine learning repository and benchmark several data-agnostic and unsupervised approaches for aggregation against ours.

The benchmark results indicate that our approach outperforms the other data-agnostic and unsupervised aggregation approaches. It is almost on par with linear regression.
\end{abstract}



\begin{keyword}
Data aggregation\sep Unsupervised machine learning \sep UCI repository \sep Regression \sep Evaluation Framework
\end{keyword}

\end{frontmatter}

\section{Introduction}\label{sc:intro}

Aggregation combines several input variables of an object into a single output score. This output should represent meaningful information about the object. To tackle this challenge, many approaches have been proposed and widely used both in research and in societal and technical decision making, ranking, and assessment applications. 

Different aggregation approaches are evaluated on how well the output score represents the input variables, according to some measures. However, even the input variables are often just an approximation of the latent object properties. On the other hand, aggregation has a goal, e.g., decision making, ranking, and assessment, which anyway abstracts from these latent object properties. This makes it hard to objectively evaluate and compare different aggregation approaches. The objective of the present paper is to suggest a way to objectively evaluate aggregation approaches.

Similar to aggregation, regression models also combine several input variables (predictors) into a single output (response). There are many different (supervised) machine learning approaches, ranging from simple linear regression to deep learning, that derive regression models from data, i.e., objects with known predictor and response variable values. However, in many practical situations, response variable can not be easily observed. The lack of this so-called ground truth data makes it impossible to use supervised regression approaches. Aggregation is then an alternative to regression, but the problem remains: how to objectively choose an appropriate aggregation approach.

To address the challenges above, we propose to evaluate and compare aggregations approaches in cases where training data is available. In these cases, we can compare the aggregation outcome with the ground truth and with an outcome proposed by regression models trained on this ground truth.

More specifically, we developed an empirical quantitative comparison of aggregation approaches by means of both external evaluation, i.e., comparison against ground truth and against what is achievable with supervised learning, and internal evaluation, i.e., comparison against information in the input variables. We compare basic aggregation functions and data-driven aggregation approaches created by unsupervised learning. 

To this end, we created a benchmark from a collection of 169 regression problems including the whole collection of regression data sets of the UCI machine learning repository (except for some data sets excluded for technical reasons). We trained the data-driven aggregation approaches on the predictors of the regression data sets (unsupervised learning). For comparison, we also trained regression models on these data sets (supervised learning). This allows for a statistical evaluation of different aggregation approaches and even their comparison against regression models. We expect that the basic aggregation models are inferior to the data-driven aggregation models (created with unsupervised learning) that, in turn, are inferior to the regression models (created with supervised learning). We will confirm this hypothesis and quantitatively assess the differences between the approaches.

In summary, the work contributes with:
\begin{enumerate}
    \item An approach for evaluation of aggregation approaches in a context of machine learning regression tasks.
    \item An implementation of this approach in a general and flexible framework for evaluation. This framework contains the benchmark data sets, the performance measures, and the implementations of the compared aggregation approaches (extensible with other approaches). 
    \item An extensive empirical comparison of several aggregation approaches.
\end{enumerate}

The remainder of the paper is structured as follows. We define aggregation and motivate the choice of aggregation operators we compare in Section~\ref{sc:aggregation}. We summarize related work concepts in Section~\ref{sc:related}. We introduce an evaluation framework in~\ref{sc:evalframevork}. In Section~\ref{sc:results} we report and discuss the results of the experimental comparison of several aggregation operators. Finally, Section~\ref{sc:conclude} concludes the research and points out directions of future work.

\section{Aggregation}\label{sc:aggregation}

In this section, we define and classify aggregation and introduce the approaches we evaluate and compare in our study.

\subsection{Definition}

Aggregation maps several input variables to a single output variable. We assume that the number of input variables is fixed, say $k$. W.l.o.g we also assume that all input and the output variables are of the unit interval $[0, 1]$. 
Formally, a function that maps the ($k$-dimensional) unit cube onto the unit interval
\begin{equation}\label{eq:agg_op}
A: [0,1]^k \mapsto [0,1] 
\end{equation}
is called an \emph{aggregation function} if it satisfies the following properties:

\textit{Monotonicity:}
\begin{equation}\label{eq:mon}
A(a_1,\ldots,a_k) \leq A(b_1,\ldots,b_k)\Leftrightarrow \bigwedge_{i \in 1\ldots k}\ a_i\leq b_i 
\end{equation}

\textit{Boundary conditions:} 
\begin{eqnarray} \label{eq:bound}
A(0,\ldots,0)=0 \wedge A(1,\ldots,1)=1 
\end{eqnarray}

A special case of aggregation is the \emph{aggregation of a singleton}, i.e., the unary operator $A:[0,1]\mapsto [0,1]$, that usually used to get a score or index for a single variable. In this paper, we do not consider aggregation of a singleton, i.e., from now on we make an assumption $k\geq 2$. 

\subsection{Basic Aggregation}

Basic aggregation functions can be classified into three main classes with specific behavior and semantics~\cite{beliakov2007aggregation}: conjunctive, disjunctive, and averaging. These classes are described below.

{\em Conjunctive} aggregation combines values like a logical \emph{AND} operator, i.e., the result of aggregation can be large if {\em all} values are large. The basic rule is that the output of aggregation is bounded from above by the lowest input. If one of the inputs equals $1$, then the output of aggregation is equal to the degree of our satisfaction with the other input variables. If any input is $0$, then the output must be $0$ as well. For example, if to obtain a driving license one has to pass both theory and driving tests, no matter how well one does in the theory test, it does not compensate for failing the driving test. From the set of basic conjunctive aggregations, we compare minimum \textbf{MIN} and product \textbf{PROD} in our evaluation.

{\em Disjunctive} aggregation combine values like a logical \emph{OR} operator, i.e., the result of aggregation can be large if at least one of the values is large. The basic rule is that the output of aggregation is bounded from below by the highest input. In contrast to conjunctive, satisfaction of any of the input variables is enough by itself. For example, when you come home both an open door and the alarm are indicators of a burglary, and either one is sufficient to raise suspicion. If both happen at the same time, they can reinforce each other, and our suspicion might become even stronger than the suspicion caused by any of these indicators by itself. From the set of basic disjunctive aggregations, we compare maximum \textbf{MAX} and sum \textbf{SUM} in our evaluation.

{\em Averaging} aggregation is also known as compensative and compromising aggregation); a high (low) value of one input variable can be compensated by a low (high) value for another one and the result will be something in between. The basic rule is that the output of aggregation is lower than the highest input and larger than the lowest input. Note that the aggregation function MIN (MAX) is at the same time conjunctive (disjunctive) and the extreme cases of an averaging aggregation. In the paper, we do not consider basic averaging aggregation functions, e.g.,  arithmetic and geometric mean (median), since their output is (almost) proportional to the output of SUM and PRODUCT, resp. 

\subsection{Data-Driven Aggregation}

Data-driven approaches need tuples of variable values to define the aggregation function. Unsupervised approaches only require tuples of input variable values while supervised approaches require tuples of input and out variable values. Once the aggregation function is learned, it can be applied to all possible input variable tuples. 

{\em Regression:} Widely used in research and applications are instances of weighted arithmetic mean. Weights usually indicate the importance of the input variables and can be set by experts or calculated from raw data~\cite{velasquez2013analysis}. In case both input and response variables are known, weights can be adjusted to fit the raw data by solving an optimization problem that minimizes an error. One basic way to solve this problem is to use linear regression~\cite{beliakov2007aggregation}. We refer to this supervised machine learning technique as \textbf{REG} and compare it with the other unsupervised techniques of basic and data-driven aggregation.

{\em Weighted quality scoring} (WQS) is a fully automated unsupervised approach based on the weighted product model for aggregation~\cite{ulan2021weighted}. Based on input tuples, it normalizes the input variables to be not correlated negatively, i.e., they have the same direction. It then calculates weights that account the variation of values of a single input variable and for the interdependence between all input variables. 

WQS was originally designed for software quality assessment. It was evaluated in the context of the defect prediction (i.e., a proxy of ground truth is a number of bugs). The authors motivated the choice of a weighted product model for aggregation as it provides a clear interpretation of aggregation output in the context of software quality assessment: in order to easily spot software artifacts with extremely bad values in a single metric, i.e., $0$ is an annihilator or so-called ``veto'' element, the aggregated quality should be poor even if only a single metric indicates that. In this paper, in addition to the original WQS approach, we compare also its weighted sum variant: normalization and weighting are the same, the only difference is in the final aggregation step which is a sum \textbf{WSM} or a product \textbf{WPM}.

\section{Related Concepts}\label{sc:related}

In this section, we relate aggregation to the concepts of data fusion, decision making, and machine learning.

\subsection{Data fusion}

The integration of information from several sources is known as data fusion. Different fusion techniques have been used in areas such as statistics, machine learning, sensor networks, robotics, and computer vision, to name a few~\cite{cocchi2019data}. The goal of fusion is to combine data from multiple sources to produce information better than would have been possible by using a single source. Improved information could mean less expensive, more accurate, or more relevant information. The goal of data aggregation is to combine data from multiple variables by reducing and abstracting it into a single one. In this sense, data aggregation is a subset of data fusion. In this paper, we restrict ourselves to data aggregation of a finite number of numerical input variables into a single output variable that represents meaningful properties of the input data. 

\subsection{Multi-criteria decision making}

Multi-Criteria Decision Making (MCDM) evaluates alternatives according to several criteria that are numerical variables. In order to choose the best alternative, one needs to aggregate the values of the criteria in some way. One popular approach is called the Multi-Attribute Utility Theory (MAUT)~\cite{von1975multi}. MAUT assigns a numerical score to each alternative, called its utility. The total utility is a function of individual utilities for the criteria. 
The rational decision-making axiom implies that one cannot prefer an alternative over another alternative if it performs better with respect to some individual utilities, but inferior with respect to the other ones. Mathematically, this means that the total utility is a monotone non-decreasing function with respect to all arguments. If we scale the utilities to $[0, 1]$, and add the boundary conditions, we obtain that total utility is an aggregation function~\cite{beliakov2007aggregation}. 

\subsection{Machine learning}

Machine learning (ML) defines models that map input to output variables using example values of the variables. Based on the learning approach, we distinguish {\em unsupervised} ML, where models are learned solely based on tuples of input variables, from {\em supervised} ML, where models are learned based on tuples of input and output variables, and from {\em feedback} ML, where models are learned incrementally based on feedback on suggested output values. The aggregation techniques WSM and WPM are examples of unsupervised ML; the aggregation technique REG is an example of supervised ML. The basic aggregation techniques are not ML examples. Feedback ML is not relevant in the context of the present paper.

The {\em UCI machine learning dataset repository}~\cite{Bache+Lichman:2013} has been widely used by the machine learning community for the empirical analysis of different ML approaches. In this paper, we use the subset of UCI for the regression task for our evaluation. We consider each data set as an aggregation problem, where both input variables and output variable (ground truth) are known. We use these data sets to develop a realistic and significant evaluation of the aggregation approaches disregarding the output data when training the data-driven aggregation approaches WSM and WPM. Using such datasets for evaluation purposes is not new. For example, a similar subset was used for an extensive experimental survey of regression methods~\cite{fernandez2019extensive}. The goal of that study was to evaluate predictive performance of supervised models. In this paper, we include a slightly larger data set (51 regression datasets were added to UCI after the date of the publication). The main difference is, however, that we evaluate basic and data-driven aggregation functions, the latter using unsupervised ML, in a dataset created for assessing regression models, using supervised ML. We add simple linear regression REG, a supervised ML approach, as a reference point. We compare aggregation functions on how good their output preserves properties of the input data (internal evaluation), and how good output agrees with the ground truth (external evaluation). To the best of our knowledge, such an empirical comparison of aggregation functions hasn't be performed before. 

Orthogonally to a classification by the availability of training data (in supervised, unsupervised, and feedback learning), ML approaches can also be classified based on their purpose. If the models are used for {\em prediction}, they are considered as black-boxes and their prediction accuracy is the foremost selection criterion of the ML approach. If they are used for {\em inference}, i.e., human knowledge gain, then their understandability is more important than accuracy for selecting an appropriate ML approach. It is well known that understandability and accuracy of the ML approaches are negatively correlated. In the present paper, we focus on prediction and, hence, evaluate the accuracy of the aggregation approaches, not their interpretability.  

{\em Dimensionality reduction} is the embedding of elements of a high-dimensional vector space in a lower-dimensional target space~\cite{van2009dimensionality}. It is an unsupervised ML approach. Implementations of dimensionality reduction include principal component analysis (PCA)~\cite{doi:10.1080/14786440109462720}, stochastic neighbor embedding (SNE)~\cite{NIPS2002_6150ccc6}, and its  $t$-distributed variant $t$-SNE~\cite{JMLR:v9:vandermaaten08a}. One could think that the special case of reducing a multi-dimensional vector space (of input variables) to just one dimension (of a single output variable) is a problem equivalent to aggregation. However, dimensionality reduction only aims at preserving the neighborhood of vectors of the original space in the reduced space. In contrast to that, aggregation assumes a latent ground truth inducing a total order in the data related to the orders induced by each input variable that is to be aggregated. Consequently, the accuracy of dimensionality reduction can be evaluated based on the observed data, i.e., the elements of a high-vector space, while the accuracy evaluation of aggregation additionally needs an explicit ground truth. Also, dimensionality reduction is a ML technique for inference while aggregation is a technique for predicting a (latent) ground truth.  

\section{Evaluation framework}\label{sc:evalframevork}
This section describes the setup used in our evaluation framework. We first present the data sets. We then provide the details for approaches we used for comparison followed by performance measures used to assess their performance.

\subsection{Data} 
We analyzed the snapshot of UCI from February 2021. We adapt the regression task that contains datasets across a wide variety of domains to build a benchmark for comparison\footnote{\url{https://archive.ics.uci.edu/ml/datasets.php?format=&task=reg}, (visited on February 22, 2021).}. The datasets included in the benchmark were chosen by the following criteria: (i) data is available and its format is numerical excluding, e.g., text and images files, (ii) the response variable is clearly defined and there are more than 10 different output values, otherwise it is potentially confused with classification, and (iii) the number of instances is greater than 50.

From the 134 available datasets, we excluded 4 since they are duplicates, i.e., identical to another regression task in UCI. Then we removed 6 because of a missing/broken link to the data and 4 since input variables are not numerical (i). We removed 26 since the response variable is not clearly defined and 30 since there are less than 10 different output values (ii). Finally, we removed 2 datasets with too few data points (iii). See \autoref{tab:excluded} in Appendix A for the detailed list of datasets with reasons for exclusion.

We selected the remaining 62 original UCI datasets that come from different domain areas: 10 from Life Sciences, 14 from Physical Sciences, 24 from CS/Engineering, 4 from Social Sciences, 7 from Business, and 3 from other fields. 

Some of them contain several sub-datasets and regression tasks and we used all of them. Therefore, the final benchmark consists of 169 datasets. See \autoref{tab:selected1} and \autoref{tab:selected2} in Appendix for a detailed list of the datasets selected for this study. 

The number of inputs for aggregation problem in these datasets differs from 2 to 373, the number of instances from 60 to 4\,208\,261, and the number of different output values differs from 14 to 72\,746. \autoref{tb:datasets_stats} summarizes some descriptive statistics for these datasets.

\begin{table}[htb]
\caption{Descriptive statistics for the datasets used in experiments.\label{tb:datasets_stats}}
\centering
\begin{tabular}{@{}lrrr@{}}
\toprule
     & \#instances &\#input variables &\#output values \tabularnewline 
\midrule
min         &  60   &  2  & 14\tabularnewline
median         &  9\,784   &  9  & 515\tabularnewline
max      &  4\,208\,261  &  373   & 72\,746\tabularnewline
\bottomrule
\end{tabular}
\end{table}

\subsection{Aggregation approaches}
Below we formally define the aggregation approaches selected for comparison (we motivated our choice in \autoref{sc:related}).

\textbf{PROD}, \textbf{MIN}, \textbf{MAX}, and \textbf{SUM} apply the respective basic aggregation functions:
\begin{eqnarray}
PROD &:=&\prod_{i=1}^k x_i\\
MIN &:=&min(x_1,\ldots,x_k)\\
MAX &:=&max(x_1,\ldots,x_k)\\
SUM &:=&\sum_{i=1}^k x_i
\end{eqnarray}
\textbf{REG} calculates a weighted arithmetic mean of the input values, \textbf{WSM} and \textbf{WPM} calculate a weighted sum and product, resp., of the normalized input values:
\begin{eqnarray}
REG &:=&\sum_{i=1}^kw_ix_i\\
WSM &:=&\sum_{i=1}^k w_i s_i(x_i) \\
WPM &:=&\prod_{i=1}^k s_i^{w_i}(x_i)
\end{eqnarray}
For the ML approaches REG, WSM, and WPM, the weights $w_i$ and the normalization functions $s_i, \text{where} \ i=1 \ldots k$, are learned from data. Let each input variable $x_1,\ldots,x_k$ and response variable $y$ have $n$ instances; we denote their $j$-th values by $x_{1j},\ldots, x_{kj},$ and $y_j$, resp. 

For REG, weights are learned using the least squares approach to minimize the sum of error squares in the training data: 
\begin{eqnarray}\label{eq:reg_learning}
 \text{minimize}\   \sum_{j=1}^n \left(\sum_{i=1}^k w_ix_{ij}-y_j \right)^2\\
 \text{subject to}\  \sum_i^k w_i = 1, \forall{i} \ w_i\geq 0 \nonumber
\end{eqnarray}

For WSM and WPM, scores and weights are learned as follows. Let $\mathds{1}$ be the \emph{indicator function} with $\mathds{1}(\mathit{cond}) = 1$ if $\mathit{cond}$ and $0$, otherwise, and let $cor$ be {\em Pearson's} coefficient of correlation.
\begin{eqnarray}\label{eq:start}
s_{i}(x_{ij})&=&\frac{1}{n}\sum_{j'=1}^{n}\mathds{1}
\begin{cases}
    (x_{ij'}\leq x_{ij}),& \text{if}\ cor(x_i,x_1)\geq 0 \\
    (x_{ij'}\geq x_{ij}),& \text{otherwise} 
\end{cases}\\
w_i&=&\frac{ w^{dep}_{i}\times w^{ent}_{i}}{\sum_{i'=1}^k  \left( w^{dep}_{i'}\times w^{ent}_{i'}\right)}\\
w^\textit{ent}_{i}&=& \frac{H_i}{\sum_{i'=1}^k H_{i'}}\\
w^{dep}_{i}&=&\frac{1-\rho_i}{k-\sum_{i'=1}^k \rho_{i'}}
\end{eqnarray}

For the entropy based weights $w^{ent}_{i}$, we define $H_i$, the {\em entropy} of the normalized variable $x_i$. Let $s_{ij} = s_{i}(x_{ij})$.
\begin{eqnarray}
H_i &=&-\sum _{j=1}^{n}p(s_{ij})\log p(s_{ij})
\end{eqnarray}
where $p(s_{ij})$ is the empirical frequency of $s_{ij}$.

For the dependency based weights $w^{dep}_{i}$, we define $\rho_i$, the {\em dependency of aggregation}  on each variable $x_i$. Therefore, let $R_i$ and $R$ be two rankings of the data points $(x_{1j}, \ldots, x_{kj})$ ascending in $r_i$ and $r$, resp., where:
\begin{eqnarray}
 r_i(x_{1j}, \ldots, x_{kj})&=&s_{ij}\\
 r(x_{1j}, \ldots, x_{kj}) &=& \frac{1}{n}\sum_{j'=1}^{n}\mathds{1}(s_{1j'} \leq s_{1j}\land \dots \land s_{kj'} \leq s_{kj})\nonumber
\end{eqnarray}
Then $\rho_i$ is the absolute value of \emph{Spearman's} rank order correlation $\rho$ of these two rankings: 
\begin{eqnarray}\label{eq:end}
\rho_i &=& \left|\rho(R_i,R)\right|
\end{eqnarray}

The Equations \ref{eq:start}--\ref{eq:end} give a self-contained definition of how the learning of the normalization functions and the weights work for WSM and WPM. However, they are better motivated and explained in depth in~\cite{ulan2021weighted}.

Note that the supervised approach REG requires the input variables $x_{i}$ and the output variable $y$ for learning the weights, cf. \autoref{eq:reg_learning}, whereas the unsupervised approaches WSM and WPM only learn from the input variables, cf. Equations \ref{eq:start}--\ref{eq:end}.

\subsection{Evaluation measures}

The choice of evaluation measures depends on the purpose of aggregation and on the availability of a ground truth. Our evaluation aims at highlighting the best aggregation for {\em prediction} assuming a (latent) ground truth. Recall that the selected datasets from the UCI repository contain both input variables and response variables. In this study, we consider response variable as an explicitly available proxy of the ground truth.  

We use the following four measures in order to compare aggregation approaches from different points of view. We evaluate aggregation approaches taking into account both external (ground truth) and internal (raw data) information. For external evaluation, we study how well aggregation output agrees with a ground truth. Therefore, we measure \emph{predictive power} and \emph{similarity}. For internal evaluation, we study how well aggregation output represents the properties of the input variables. Therefore, we measure \emph{consensus} and \emph{sensitivity}.

\textbf{Predictive power.} We measure the correlation between aggregation output and ground truth to assess the ordering, relative spacing, and possible functional dependency. We use \emph{Spearman's rho}~\cite{spearman1904general} as a correlation coefficient to measure the pairwise degree of association between the values, i.e., how well a monotonic function describes the relationship. High values indicate a strong predictive power.

\textbf{Similarity.} Moreover, we rank the data points according to the aggregation output and to the ground truth in acceding order. We measure a distance between the two rankings based on the \emph{Kendall's tau distance} to assess the number of pairwise disagreements between two rankings~\cite{kendall1948rank}. It corresponds to the number of transpositions that bubble sort requires to turn one ranking into the other. Low values indicate a high similarity. 

\textbf{Consensus.} We rank the data points according to aggregation output and according to each input variable in acceding order. We use the \emph{Kemeny distance}~\cite{dwork2001rank}, i.e., the sum of the $k$ Kendall tau distances between aggregation output and input rankings, to assess a consensus between the aggregation output and the input variables. Low values indicate a strong consensus. 

\textbf{Sensitivity.} Finally, we measure how well the aggregation preserves a variety of the input data. We use the \emph{sensitivity ratio}~\cite{ulan2021cop}, i.e., the ratio between unique aggregation outputs and the unique tuples in the raw data of input variables. High values indicate a strong sensitivity. 

\section{Experiments and discussion}\label{sc:results}

\subsection{Experimental design}
We assess the basic aggregation functions PROD and MIN (conjunctive), MAX and SUM (disjunctive), and the data-driven aggregation functions WSM and WPM (unsupervised). The averaging aggregation REG (supervised) is a baseline reference for the basic and the unsupervised models. Since the unsupervised aggregation approaches do not use the ground truth information to build the prediction, they are not expected to perform better than the supervised model REG trained on this information in the external evaluation.

It is a well-known fact in machine learning that variables with large value ranges dominate the outcome while those with smaller value ranges get marginalized. Also in aggregation, an input variable with larger value ranges could influence the aggregation output. To have a fair comparison of the aggregation approaches, we have normalized the values of input variables to $[0,1]$ using min-max scaling.

All aggregation approaches compared in this study use the same data sets (see \autoref{tab:selected1} and \autoref{tab:selected2}), the same input variables, and the same ground truth. We did not consider categorical variables as inputs as well as their artificial representations such as dummy variables. We removed the instances with missing values for either the response variable or input variables. We also removed the identification variables, e.g., time, and id, since they should not contribute to the aggregation. We implemented all algorithms and statistical analyses in \emph{R}.\footnote{The R Project for Statistical Computing, \url{https://www.r-project.org}} We provide all \emph{R} scripts that are used to conduct the experiments in a replication package downloadable from \url{https://doi.org/10.5281/zenodo.5091819}.

\subsection{Summary of results}
We run aggregation approaches over 169 datasets. We report the 25 percentile, the median, and the 75 percentile values of the distributions of \emph{Spearman's rho}, \emph{Kendall's tau distance}, \emph{Kemeny distance}, and \emph{Sensitivity ratio} for each aggregation over the all datasets. The boxplots in \autoref{fig:res_spearman}, \autoref{fig:res_kendall}, \autoref{fig:res_kemeny}, \autoref{fig:res_sensitivity} visualize the different aggregation approaches and the distribution of their performance in the evaluation measures.

\subsubsection{External evaluation.}
\begin{figure*}
\begin{multicols}{2}
    \includegraphics[width=\linewidth]{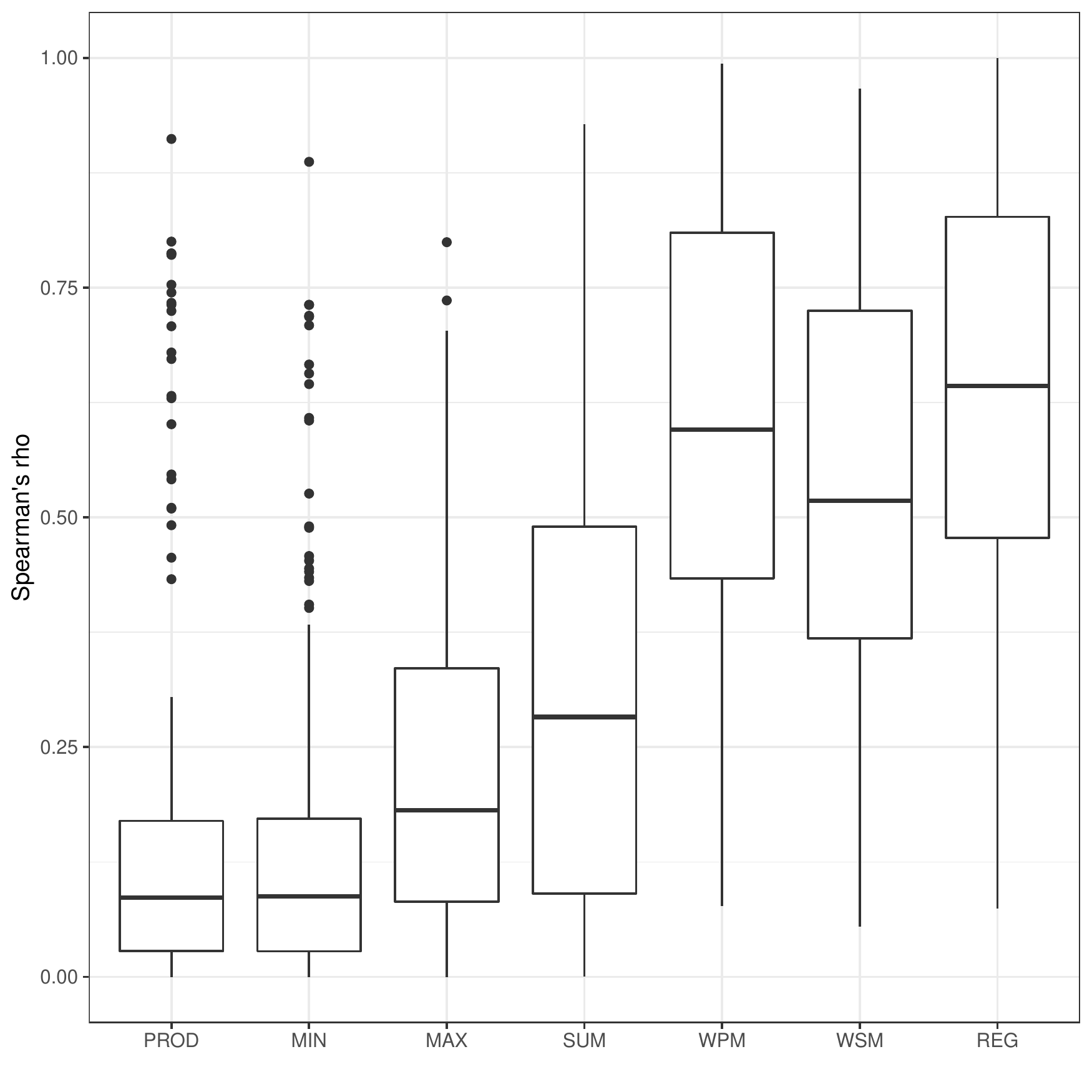}\par 
     \caption{Performance comparison in terms of \emph{Predictive power}.} 
    \label{fig:res_spearman} 
    \includegraphics[width=\linewidth]{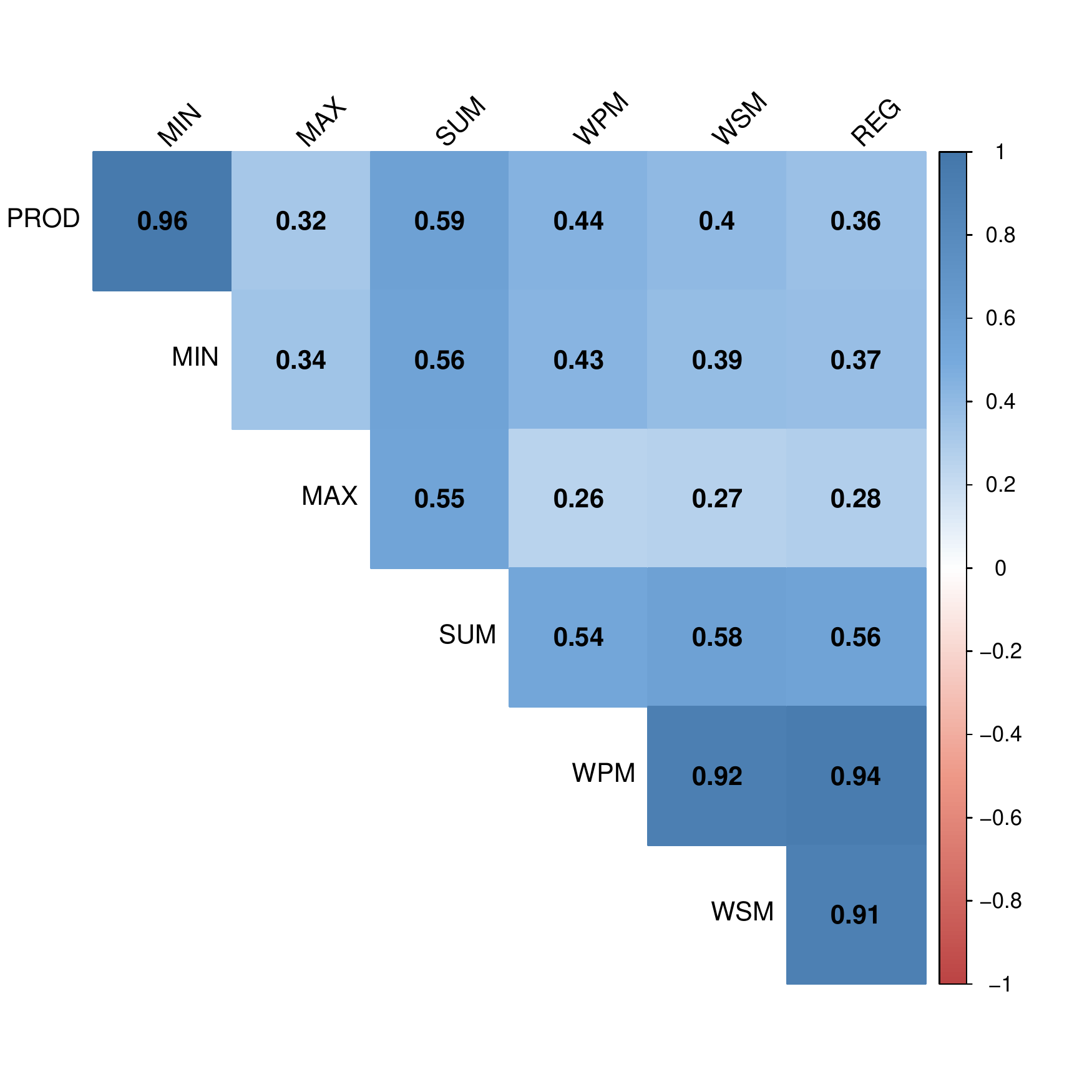}\par 
     \caption{Correlation matrix for \emph{Predictive power}.} 
    \label{fig:corplot_spearman} 
    \end{multicols}
\begin{multicols}{2}
    \includegraphics[width=\linewidth]{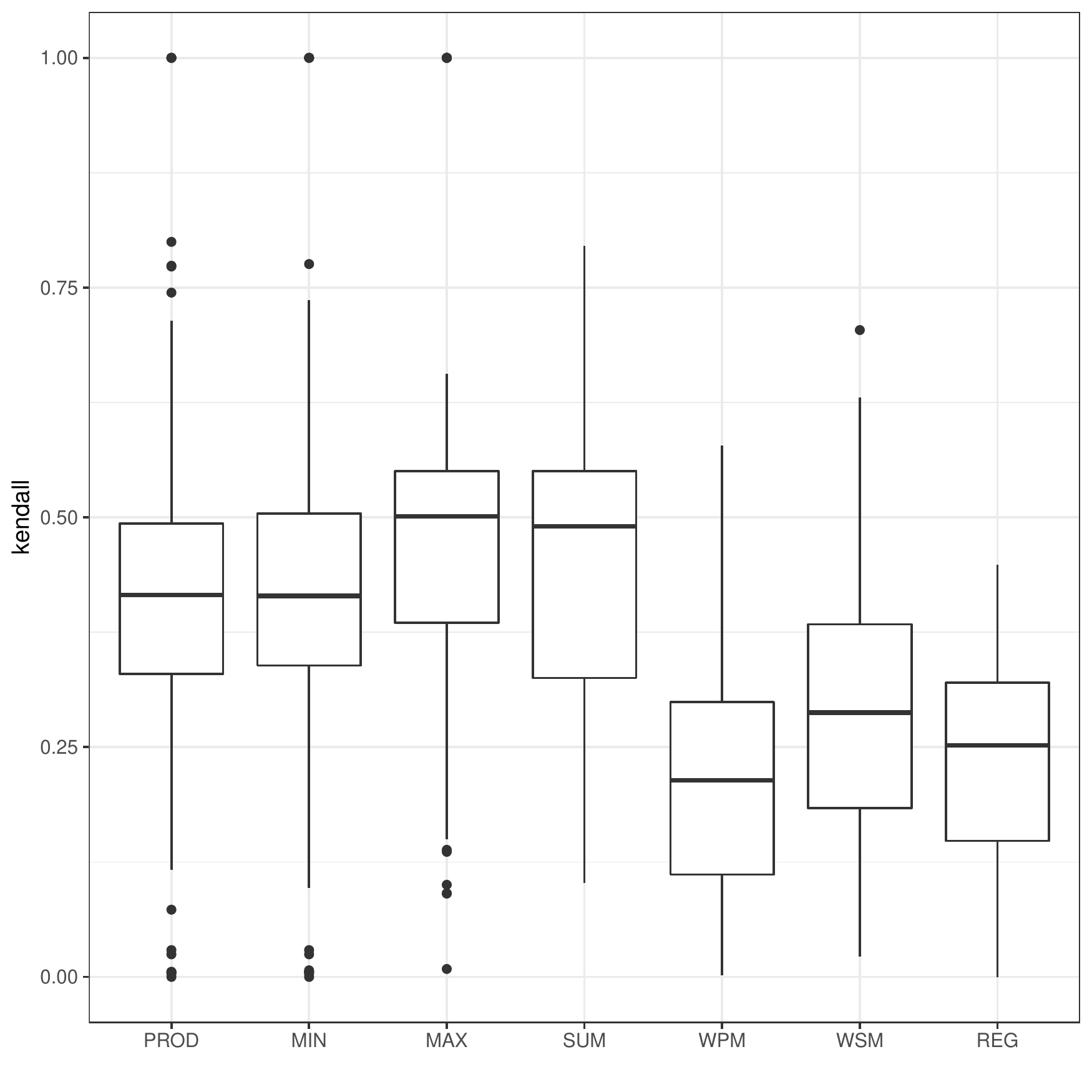}\par
     \caption{Performance comparison in terms of \emph{Similarity}.} 
    \label{fig:res_kendall} 
    \includegraphics[width=\linewidth]{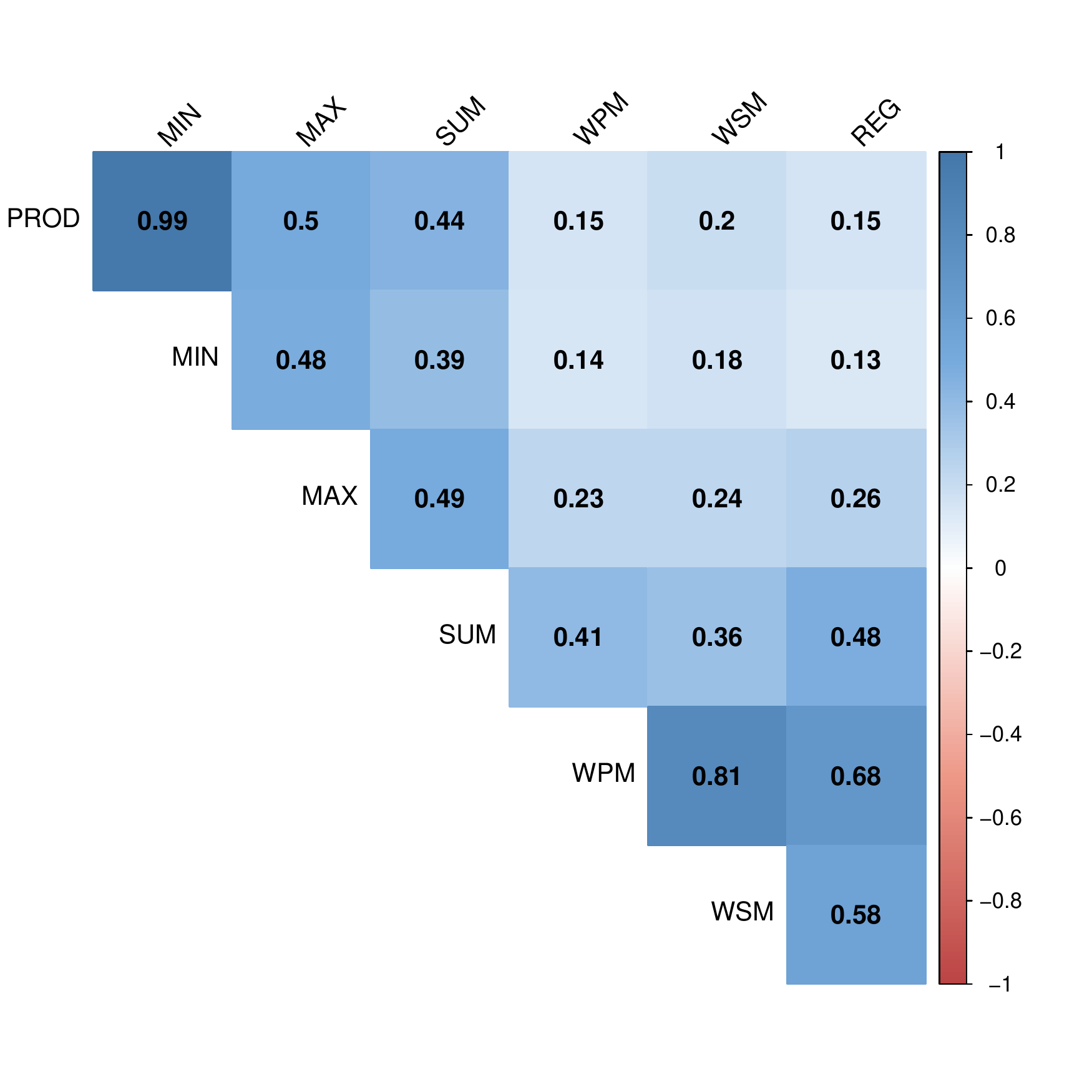}\par
     \caption{Correlation matrix for \emph{Similarity}.} 
    \label{fig:corplot_kendall} 
\end{multicols}
\end{figure*}

\autoref{fig:res_spearman} shows the performance comparisons in terms of \emph{Predictive power}. Recall that high values indicate good performance. We observe that all basic aggregation approaches perform worse than REG aggregation and the weighted scoring approaches WPM and WSM. Moreover, REG aggregation is (slightly) better than the WSM and WPM approaches. It is not a surprise, since the REG aggregation model was defined with a supervised ML approach. 

\autoref{fig:corplot_spearman} shows pairwise correlation between values of \emph{Predictive power} calculated for each approach on each dataset. We observe a very strong positive correlation between corresponding results for the data-driven approaches REG, WSM, and WPM. We also observe the same for the basic PROD and MIN approaches. Moreover, results for SUM have a high correlation with results for all other approaches. 

\autoref{fig:res_kendall} shows the performance comparisons in terms of \emph{Similarity}.  Recall that low values indicate good performance. We observe that all basic aggregation approaches perform worse than the data-driven approaches REG, WPM, and WSM. Moreover, WPM aggregation is (slightly) better than both WSM and REG.

\autoref{fig:corplot_kendall} shows pairwise correlation between values of \emph{Similarity} calculated for each approach on each dataset. We observe a very strong positive correlation between similarity results for PROD and MIN, and the same for WPM and WSM. The results for REG have high, moderate, or even low correlation with results for other approaches. 

\subsubsection{Internal evaluation.}

\begin{figure*}
\begin{multicols}{2}
    \includegraphics[width=\linewidth]{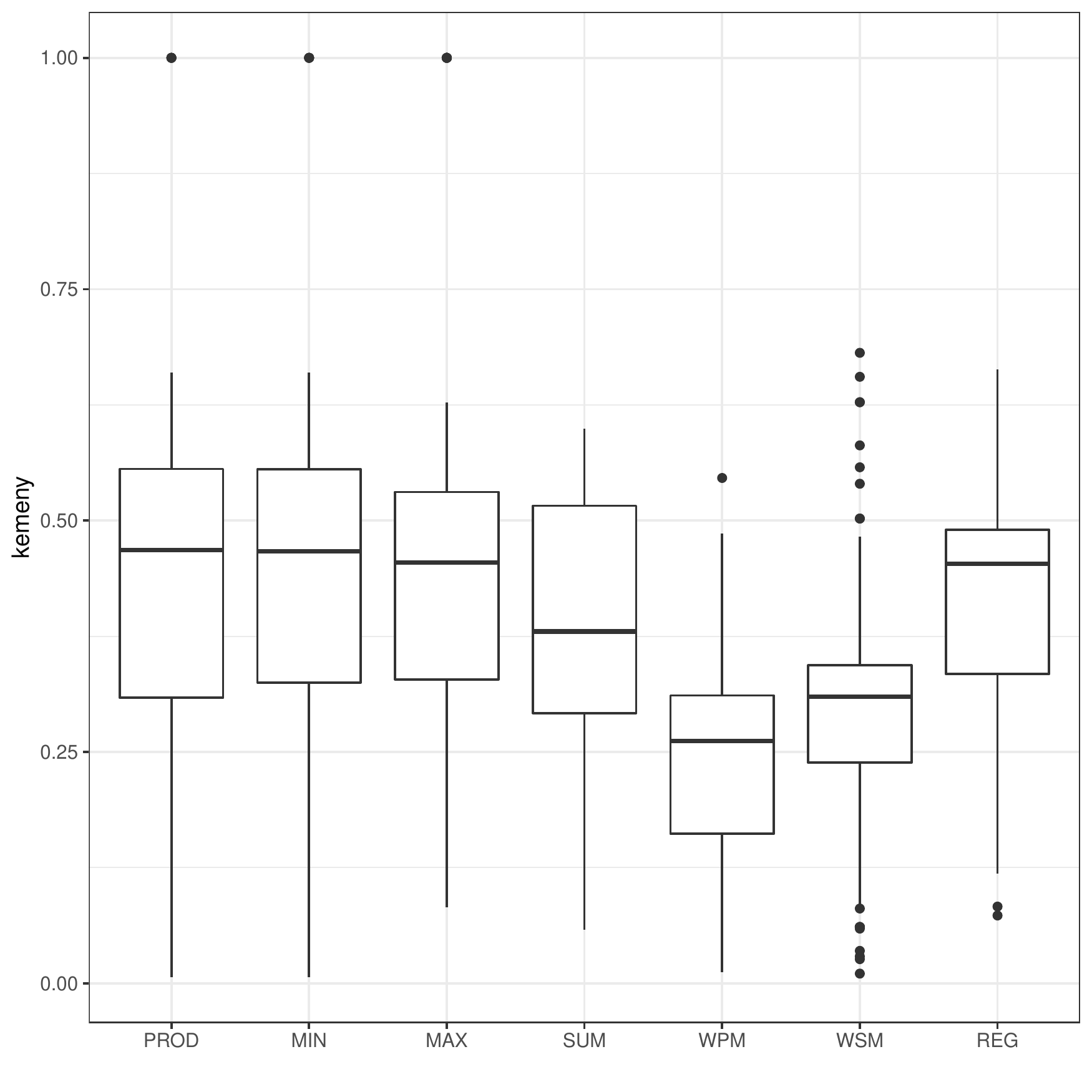}\par 
     \caption{Performance comparison in terms of \emph{Consensus}.} 
    \label{fig:res_kemeny} 
    \includegraphics[width=\linewidth]{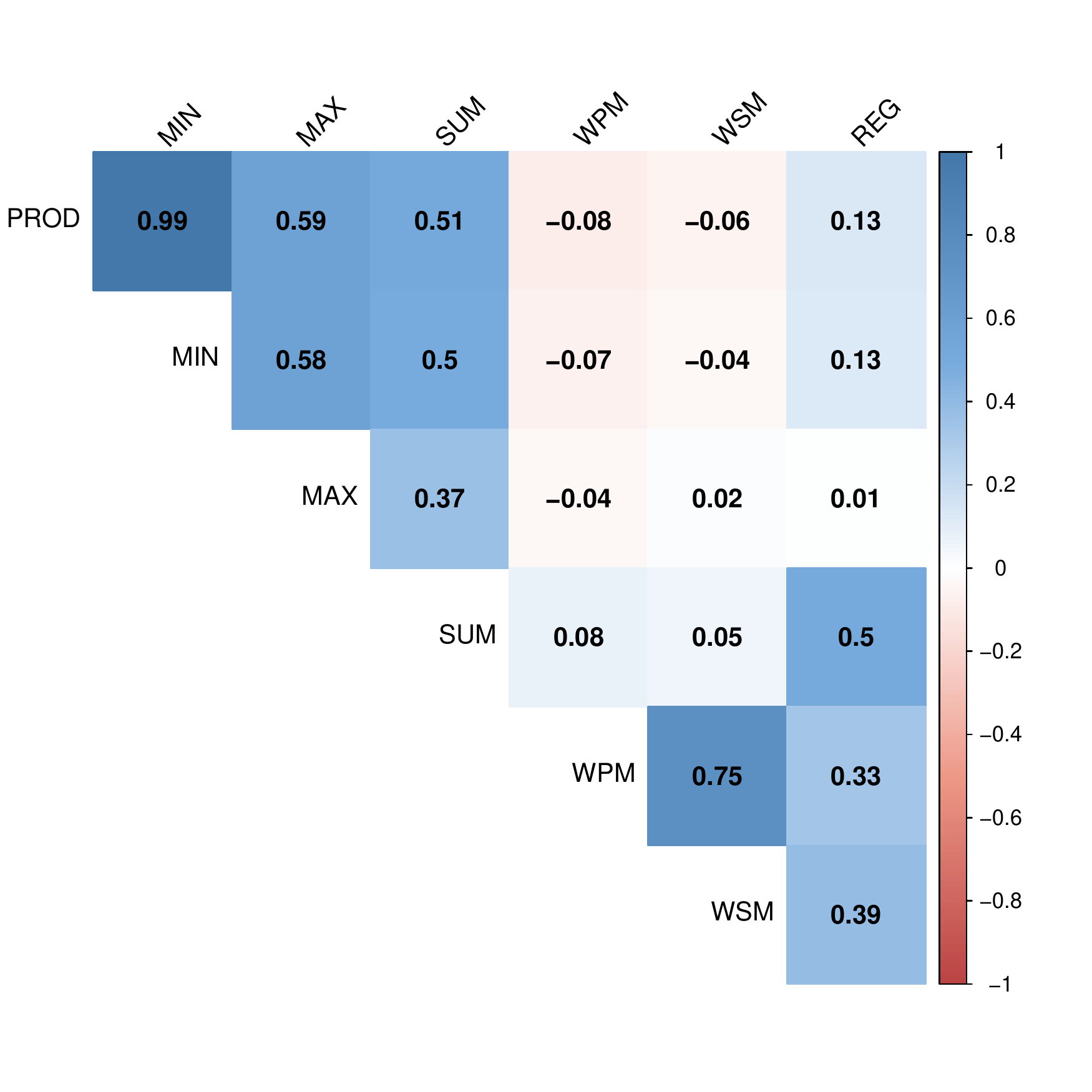}\par 
     \caption{Correlation matrix for \emph{Consensus}.} 
    \label{fig:corplot_kemeny} 
    \end{multicols}
\begin{multicols}{2}
    \includegraphics[width=\linewidth]{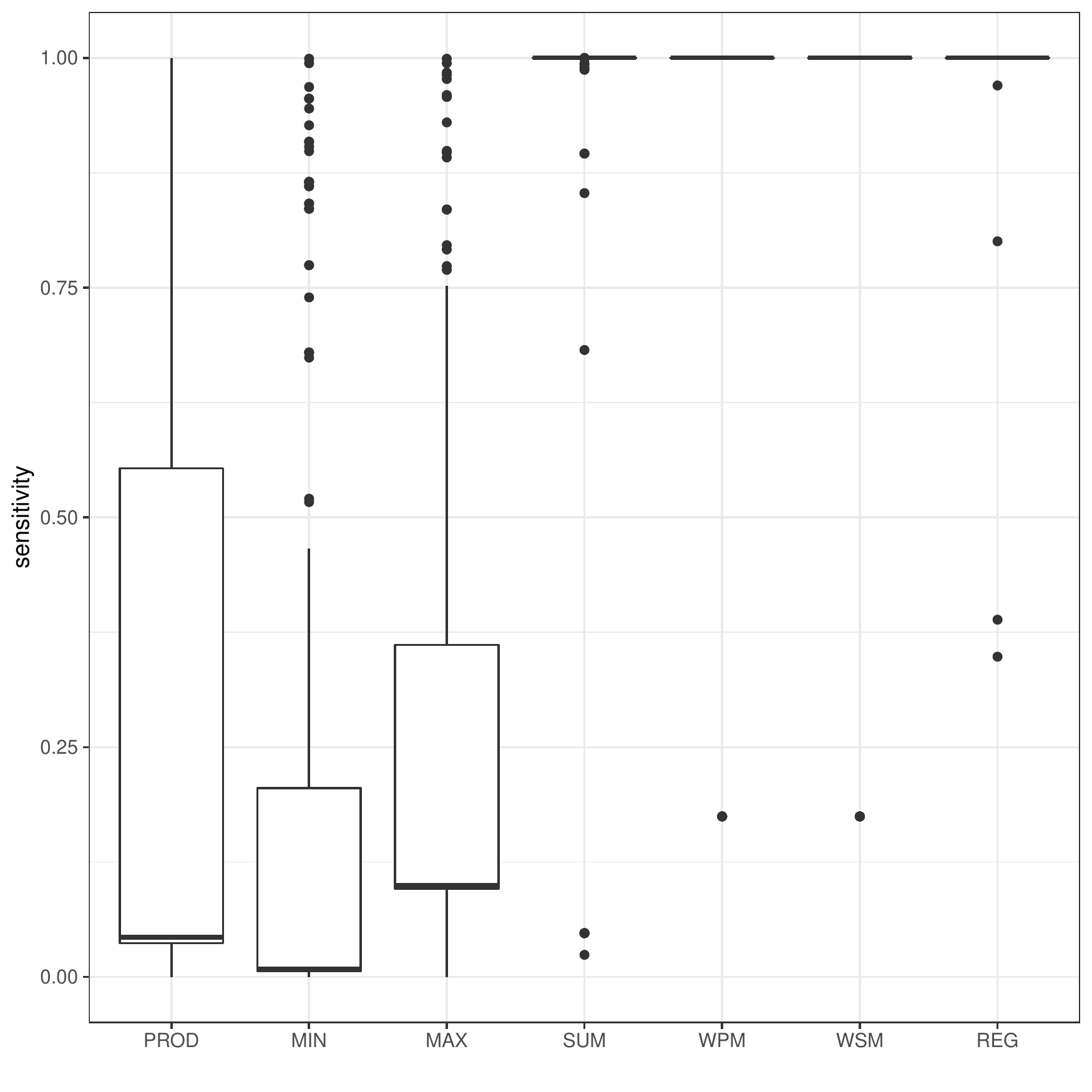}\par
     \caption{Performance comparison in terms of \emph{Sensitivity}.} 
    \label{fig:res_sensitivity} 
    \includegraphics[width=\linewidth]{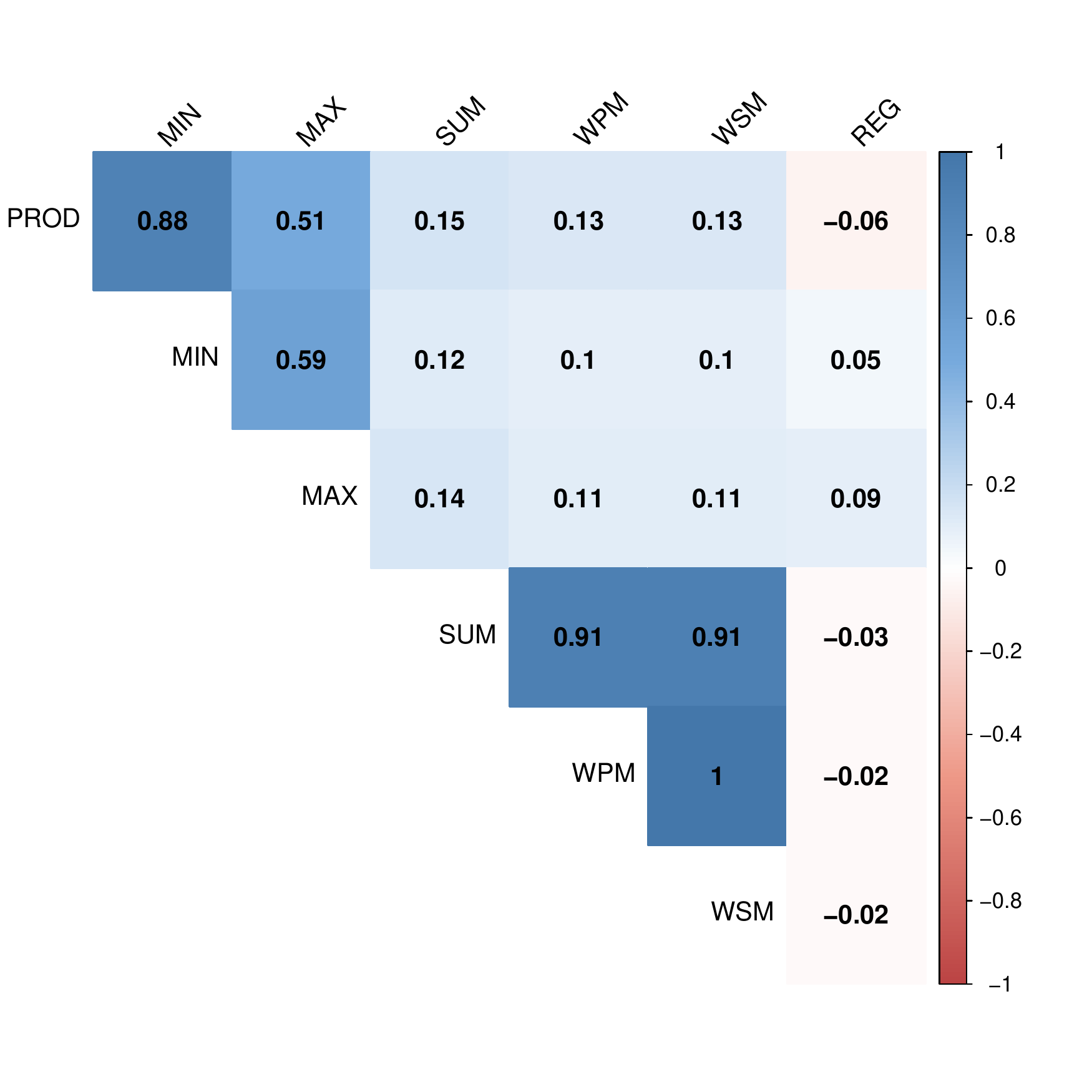}\par
     \caption{Correlation matrix for \emph{Sensitivity}.} 
    \label{fig:corplot_sensitivity} 
\end{multicols}
\end{figure*}

\autoref{fig:res_kemeny} shows the performance comparisons in terms of \emph{Consensus}. Recall that low values indicate good performance. We observe that all basic and REG aggregation approaches perform worse than the weighted scoring approaches WPM and WSM. WPM, in turn, is slightly better than  WSM. 

\autoref{fig:corplot_kemeny} shows pairwise correlation between the \emph{Consensus} results calculated for each approach on each dataset. We observe a very strong positive correlation between the results for PROD and MIN, and a strong correlation between corresponding results for  WPM and WSM. The results for REG have a high, moderate, or even low correlation with results for other approaches.  
\autoref{fig:res_sensitivity} shows the performance comparisons in terms of \emph{Sensitivity}. Recall that high values indicate good performance. We observe that REG, WPM, WSM, and SUM perform equally well, while the other basic aggregation approaches perform significantly worse. 

\autoref{fig:corplot_sensitivity} shows pairwise correlation between values of \emph{Sensitivity} calculated for each approach on each dataset. We observe a very strong positive correlation between the results for PROD and MIN, and a strong correlation between the results for WPM and WSM. We observe a high or moderate correlation between the results for basic aggregation approaches, and the same for the results between REG, and the SUM, WPM, and WSM approaches. We also observe an extremely low correlation between the results for data-driven and basic aggregation approaches (except correlation between the results for REG and SUM). Note that, it does not mean that approaches have completely different sensitivity results. Instead, this is an effect of the results of sensitivity for data-driven approaches that have too few different values--for almost all datasets, the sensitivity is one and the correlation results are dominated by the few outliers in different datasets.

\subsubsection{Summary and Discussion}

\autoref{tb:eval_all} summarizes the median values of the performance measures over all datasets. The Beijing Multi-Site Air Quality Data datasets represent almost half (72 out of 169 datasets). Since this fact might bias the results, we also calculated the median values for performance measures discarding these datasets. The results are presented in parentheses.    

\begin{table}[htb]
\caption{Median values of the performance measures.}\label{tb:eval_all}
\centering
\resizebox{\columnwidth}{!}{%
\begin{tabular}{|c|c|c|c|c|}
\hline 
     & Predictive power& Similarity & Consensus & Sensitivity \\
     Aggregation & in Spearman's  rho & in Kendall tau distance & in Kemeny distance & in Sensitivity Ratio \\
\hline 
PROD& 0.09 (0.15) & 0.41 (0.41) & 0.47 (0.4) & 0.04 (0.51)\\
\hline 
MIN& 0.09 (0.15) & 0.41 (0.42) & 0.47 (0.42) & 0.01 (0.11)\\
\hline 
MAX& 0.18 (0.19) & 0.5 (0.45) & 0.45 (0.41) & 0.1 (0.29)\\
\hline
SUM& 0.28 (0.31) & 0.49 (0.39) & 0.38 (0.32) & 1 (1)\\
\hline
WPM& 0.59 (0.77) & 0.21 (0.13) & 0.26 (0.22) & 1 (1)\\
\hline
WSM& 0.52 (0.68) & 0.29 (0.24) & 0.31 (0.32) & 1 (1)\\
\hline
REG& 0.64 (0.79) & 0.25 (0.18) & 0.45 (0.39) & 1 (1)\\
\hline
\end{tabular}
}
\end{table}

We observe that the REG and WPM approaches perform (slightly) better than WSM in the external evaluation. Moreover, REG performs slightly better than WPM in terms of \emph{predictive power}. However, it is the opposite in terms of \emph{similarity}. WPM and WSM perform better than or equally well as the other approaches in the internal evaluation. Moreover, WPM performs better than WSM in terms of \emph{Similarity}.  SUM from the basic aggregations performs equally good as the data-driven approaches REG, WPM, WSM approaches in terms of \emph{sensitivity}.     

Studying the pairwise correlations between performances in the different evaluation measures for each dataset, leads to the following observations. The data-driven approaches REG, WPM, and WSM are closely associated with each other in the external evaluation measures, i.e., their respective results of \emph{Predictive power} and \emph{Similarity} are highly correlated. This means that it is not very likely that in-depth studies find properties of the datasets that favor either of the methods. 

SUM is associated better than the other basic aggregations with REG, WPM, and WSM in terms of external evaluation (i.e., moderate correlation).

The unsupervised approaches WPM and WSM are closely associated with each other in the internal evaluation measures, i.e., their respective results of \emph{Consensus} and \emph{Sensitivity} are highly correlated. REG differs from other approaches. However, associated with SUM in terms of \emph{Consensus}. We conclude that SUM from basic aggregation functions is quite competitive with REG aggregation in the internal evaluation. 

We also conclude that in terms of their overall performance (including both internal and external evaluation) the unsupervised, weighted scoring approaches WSM and WPM are on par with the supervised REG aggregation.

The experiments were performed on more than a hundred datasets with different sizes from different domains. However, this specific sample might be a {\em threat to external validity}. Further replications of this study on other datasets are necessary to confirm the generalization of the above conclusions.

\section{Conclusion and future work }\label{sc:conclude}
We evaluated aggregation approaches on a benchmark of 169 regression datasets of the UCI machine learning repository from different sciences and application domains. We empirically compared six unsupervised aggregation approaches, more specifically, four basic aggregation functions PROD, MIN, MAX, and SUM, and two data-driven aggregation approaches WSM and WPM. As a point of reference, we also assessed a supervised approach REG, i.e., a weighted sum aggregation with weights defined using linear regression. 

The aggregation approaches were evaluated externally, i.e., we compared aggregation outputs against the response variable as a proxy of a (in aggregation usually latent) ground truth. Then they were evaluated internally, i.e., we compared aggregations output against the values of the input variables. 

For external evaluation, the supervised aggregation REG achieves the best results closely followed by the unsupervised approaches WPM and WSM. For internal evaluation, WPM achieves the best results closely followed by WSM and SUM. We conclude that basic aggregation functions can be significantly improved by unsupervised ML-based approaches, that the latter are quite competitive to linear regression, a (simple) supervised learning approach. As a consequence, these data-driven aggregation approaches are well-suited for prediction tasks when ground truth is not available for training a regression model. This confirms the results of \cite{ulan2021weighted} and generalizes them from the field of bug prediction to a wider range of scientific and application fields.

For the evaluation, we developed a reusable and extensible evaluation framework, i.e., new aggregation approaches as well as new performance measures can be easily plugged in. This way other researchers are able to easily compare their aggregation approaches against the ones studied in the present paper using our evaluation framework. Provided that our benchmark and selection of aggregation approaches is considered representative enough, the proposed evaluation framework can serve as a practical guideline for selecting an appropriate aggregation approach and stimulate research in the field of aggregation, highlighting the current champion approaches and forgetting about the poorly performing ones. 

In the future, we plan to develop a visualization tool for easy comparison. Also, we plan to extend the framework by considering classification and clustering tasks using aggregation. One possible way is to apply the \emph{logit function} to the aggregation output, and then evaluate its performance as a classifier. Another interesting direction for future research is to evaluate not only the aggregation output for prediction, but its parameters, such as weights, for inference tasks.

\bibliography{evalframe.bib}

\begin{table*}[htb]
\centering
    \caption{Appendix - List of UCI regression task data sets that were excluded.} \label{tab:excluded}
    \begin{adjustbox}{width=0.85\textwidth}
\begin{filecontents*}{excluded_data.csv}
Excluded data set, reason to exclude
AI4I 2020 Predictive Maintenance,unique(output) = 2
Alcohol QCM Sensor,unique(output) = 2
Algerian Forest Fires,unique(output) = 2
Amazon Access Samples,unique(output) = 2
Apartment for rent classified, dataset is missing
Behavior of the urban traffic of the city of Sao Paulo in Brazil, response variable is not clearly defined
Bar Crawl: Detecting Heavy Drinking, response variable is not clearly defined
Bone marrow transplant: children,unique(output) = 2
Breast Cancer Wisconsin (Prognostic),unique(output) = 2
Carbon Nanotubes, response variable is not clearly defined
Cargo 2000 Freight Tracking and Tracing,unique(output) = 3
Challenger USA Space Shuttle O-Ring, too few instances
clickstream data for online shopping, response variable is not clearly defined
CNNpred: CNN-based stock market prediction using a diverse set of variables, response variable is not clearly defined
Condition monitoring of hydraulic systems,unique(output) = 2/3/4
Container Crane Controller, too few instances  
Cuff-Less Blood Pressure Estimation, response variable is not clearly defined  
Demand Forecasting for a store, dataset is missing
DrivFace,unique(output) = 4
Drug Review (Druglib.com), text input
Drug Review (Drugs.com), text input    
Dynamic Features of VirusShare Executables, response variable is not clearly defined
Early biomarkers of Parkinsons disease based on natural connected speech, dataset is missing
Educational Process Mining (EPM): A Learning Analytics, response variable is not clearly defined 
EEG Steady-State Visual Evoked Potential Signals, response variable is not clearly defined
ElectricityLoadDiagrams20112014, response variable is not clearly defined  
Estimation of obesity levels based on eating habits and physical condition,unique(output) = 7
Fertility,unique(output) = 2
Gas Sensor Array Drift Dataset at Different Concentrations,unique(output) = 7
Gas sensor array exposed to turbulent gas mixtures, response variable is not clearly defined
Gas sensor array temperature modulation, response variable is not clearly defined
Gas sensor array under flow modulation,unique(output) = 4
Geo-Magnetic field and WLAN dataset for indoor localisation from wristband and smartphone, response variable is not clearly defined
GNFUV Unmanned Surface Vehicles Sensor Data, response variable is not clearly defined
GNFUV Unmanned Surface Vehicles Sensor Data Set 2, response variable is not clearly defined  
Heart failure clinical records,unique(output) = 2
Hungarian Chickenpox Cases, response variable is not clearly defined
IIWA14-R820-Gazebo-Dataset-10Trajectories,unique(output) = 2
Improved Spiral Test Using Digitized Graphics Tablet for Monitoring Parkinson’s Disease, dataset is missing
Incident management process enriched event log, response variable is not clearly defined
Insurance Company Benchmark (COIL 2000),unique(output)=3
Iranian Churn, dataset is missing
KDC-4007 dataset Collection,unique(output)=8
KDD Cup 1998, response variable is not clearly defined
Las Vegas Strip,unique(output)=5
News Popularity in Multiple Social Media Platforms, text input  
NoisyOffice, images input
Online Retail II, response variable is not clearly defined
Open University Learning Analytics, response variable is not clearly defined    
Paper Reviews,unique(output)=5
Parking Birmingham Data Set, response variable is not clearly defined
Parkinson Disease Spiral Drawings Using Digitized Graphics Tablet,unique(output)=3
Pedestrian in Traffic, response variable is not clearly defined
PPG-DaLiA Data Set, response variable is not clearly defined
QSAR fish bioconcentration factor (BCF), response variable is not clearly defined
Query Analytics Workloads, response variable is not clearly defined
Simulated data for survival modelling, response variable is not clearly defined
SkillCraft1 Master Table,unique(output)=7
Solar Flare,unique(output)=3
South German Credit,unique(output)=2
South German Credit (UPDATE),unique(output)=2
Tamilnadu Electricity Board Hourly Readings,unique(output)=2
Tennis Major Tournament Match Statistics,unique(output)=2
Twin gas sensor arrays,unique(output)=4
UJIIndoorLoc-Mag,unique(output)=1
WESAD (Wearable Stress and Affect Detection),dataset is missing
wiki4HE,unique(output)=7
Wine Quality,unique(output)=7 / 6 
\end{filecontents*}
\csvautotabular{excluded_data.csv}
     \end{adjustbox}
\end{table*}

\begin{table*}[htb]
\centering
    \caption{Appendix - List of UCI regression task data sets that were selected.} \label{tab:selected1}
    \begin{adjustbox}{width=0.85\textwidth}
\begin{filecontents*}{selected_data_1.csv}
name,Original UCI name,tasks,instances,inputs
3Droad,3D Road Network (North Jutland, Denmark),1,434874,3
airquality,Air Quality,5,9357,8
airfoil,Airfoil Self-Noise,1,1503,5
appliances,Appliances energy prediction,1,19735,27
autompg,Auto MPG,1,392,7
automobile,Automobile,1,159,13
PRSA\_Aotizhongxin,Beijing Multi-Site Air-Quality Data,6,31815,5
PRSA\_Changping,Beijing Multi-Site Air-Quality Data,6,32681,5
PRSA\_Dingling,Beijing Multi-Site Air-Quality Data,6,31306,5
PRSA\_Dongsi,Beijing Multi-Site Air-Quality Data,6,30338,5
PRSA\_Guanyuan,Beijing Multi-Site Air-Quality Data,6,32263,5
PRSA\_Gucheng,Beijing Multi-Site Air-Quality Data,6,32504,5
PRSA\_Huairou,Beijing Multi-Site Air-Quality Data,6,31708,5
PRSA\_Nongzhanguan,Beijing Multi-Site Air-Quality Data,6,33114,5
PRSA\_Shunyi,Beijing Multi-Site Air-Quality Data,6,30194,5
PRSA\_Tiantan,Beijing Multi-Site Air-Quality Data,6,32843,5
PRSA\_Wanliu,Beijing Multi-Site Air-Quality Data,6,30634,5
PRSA\_Wanshouxigong,Beijing Multi-Site Air-Quality Data,6,32768,5
beijing\_pm25,Beijing PM2.5 Data,1,41757,6
bias,Bias correction of numerical prediction model temperature forecast,2,7588,21
bike\_day,Bike Sharing Dataset,1,731,6
bike\_hour,Bike Sharing Dataset,1,17379,6
blogfeedback,BlogFeedback,1,52397,11
TomsHardware,Buzz in social media,1,28179,96
Twitter,Buzz in social media,1,583250,71
ccpp,Combined Cycle Power Plant,1,9568,4
communities,Communities and Crime,1,1994,122
comm\_unnorm,Communities and Crime Unnormalized,1,2215,124
machine,Computer Hardware,1,209,6
concrete,Concrete Compressive Strength,1,1030,8
slump,Concrete Slump Test,3,104,7
CBM,Condition Based Maintenance of Naval Propulsion Plants,1,11934,16
CSM,CSM (Conventional and Social Media Movies) Dataset 2014 and 2015,1,231,9
dailydemand,Daily Demand Forecasting Orders,1,60,12
elect\_grid,Electrical Grid Stability Simulated Data,1,10000,13
energy\_efficiency,Energy efficiency,2,768,7
facebook\_comment,Facebook Comment Volume Dataset,1,40949,52
facebook\_metrics,Facebook metrics,1,495,17
forestfires,Forest Fires,1,517,8
ethylene\_CO,Gas sensor array under dynamic gas mixtures,1,4208261,17
ethylene\_methane,Gas sensor array under dynamic gas mixtures,1,4178504,17
gt\_2011,Gas Turbine CO and NOx Emission Data Set,2,7411,9
gt\_2012,Gas Turbine CO and NOx Emission Data Set,2,7628,9
gt\_2013,Gas Turbine CO and NOx Emission Data Set,2,7152,9
\end{filecontents*}
\csvautotabular{selected_data_1.csv}
     \end{adjustbox}
\end{table*}

\begin{table*}[htb]
\centering
    \caption{Appendix - List of UCI regression task data sets that were selected. Cont. }\label{tab:selected2}
    \begin{adjustbox}{width=0.85\textwidth}
\begin{filecontents*}{selected_data_2.csv}
name,Original UCI name,tasks,instances,inputs
gt\_2014,Gas Turbine CO and NOx Emission Data Set,2,7158,9
gt\_2015,Gas Turbine CO and NOx Emission Data Set,2,7384,9
default\_features,Geographical Original of Music,2,1059,68
default\_chromatic,Geographical Original of Music,2,1059,116
gps,GPS Trajectories,1,163,3
greenhouse,Greenhouse Gas Observing Network,1,955167,14
household,Individual household electric power consumption,1,2049280,6
akbilgic,ISTANBUL STOCK EXCHANGE,1,536,8
RN\_Undirected,KEGG Metabolic Reaction Network (Undirected),1,64608,27
RN\_Directed,KEGG Metabolic Relation Network (Directed),1,53413,19
metro,Metro Interstate Traffic Volume,1,48204,4
OnlineNewsPopularity,Online News Popularity,1,39644,44
transcoding\_mesurment,Online Video Characteristics and Transcoding Time Dataset,1,68784,6
optical\_interconnection,Optical Interconnection Network,1,640,4
Parkinson\_multiple,Parkinson Speech Dataset with Multiple Types of Sound Recordings,1,1040,26
parkinsons,Parkinsons Telemonitoring,2,5875,7
CASP,Physicochemical Properties of Protein Tertiary Structure,1,45730,9
fivecities,PM2.5 Data of Five Chinese Cities,1,49579,7
productivity,Productivity Prediction of Garment Employees,1,691,8
qsar\_aquatic\_toxicity,QSAR aquatic toxicity,1,546,8
qsar\_bioconcentration,QSAR Bioconcentration classes dataset,1,779,9
fish\_toxicity,QSAR fish toxicity,1,908,6
real\_estate,Real estate valuation data set,1,414,5
ElectionData,Real-time Election Results: Portugal 2019,1,21643,7
slice\_localization,Relative location of CT slices on axial axis,1,53500,373
residential\_building,Residential Building Data Set,2,372,7
SeoulBike,Seoul Bike Sharing Demand,1,760,9
servo,Servo,1,167,2
sgemm\_product,SGEMM GPU kernel performance,1,241600,10
sml1,SML2010,1,2764,16
sml2,SML2010,1,1373,16
stock\_portfolio,Stock portfolio performance,6,315,6
student\_mat,Student Performance,1,395,13
student\_por,Student Performance,1,649,13
superconduct,Superconductivty Data,1,21263,9
UJIndoorLoc,UJIIndoorLoc,2,19937,125
WEC\_Adelaide,Wave Energy Converters,1,71999,48
WEC\_Perth,Wave Energy Converters,1,72000,48
WEC\_Sydney,Wave Energy Converters,1,72000,48
WEC\_ Tasmania,Wave Energy Converters,1,72000,48
yacht\_hydrodynamics,Yacht Hydrodynamics,1,308,6
YearPredictionMSD,YearPredictionMSD,1,515345,90
\end{filecontents*}
\csvautotabular{selected_data_2.csv}
     \end{adjustbox}
\end{table*}

\end{document}